\documentclass[letterpaper, 10 pt, journal, twoside]{IEEEtran}

\usepackage[T1]{fontenc}
\usepackage{cite}
\usepackage{amssymb,amsfonts}
\usepackage{blindtext}
\makeatletter
\let\NAT@parse\undefined
\makeatother
\usepackage{hyperref}
\usepackage{graphicx}
\usepackage{textcomp}
\usepackage{mathtools}
\usepackage{changes}
\usepackage{algorithm}
\usepackage{xcolor}
\usepackage{color, soul}
\usepackage{amsmath}
\usepackage{booktabs}
\usepackage{multirow}
\usepackage{xstring}
\usepackage{hhline}
\usepackage{kotex}
\usepackage{comment}

\usepackage{tikz}
\usepackage{cleveref}
\usepackage{lipsum}
\usepackage{bm}
\usepackage{array}
\usepackage{eqparbox}

\usepackage{subcaption}

\usepackage{siunitx}
\usepackage{physics}

\usepackage[noend]{algorithmic}

\usepackage{etoolbox}  % patch def of algorithmic environment
\makeatletter
\patchcmd{\algorithmic}{\addtolength{\ALC@tlm}{\leftmargin} }{\addtolength{\ALC@tlm}{\leftmargin}}{}{}
\makeatother

\crefformat{section}{\S#2#1#3} % see manual of cleveref, section 8.2.1
\crefformat{subsection}{\S#2#1#3}
\crefformat{subsubsection}{\S#2#1#3}

\floatname{algorithm}{Algorithm}

\definecolor{rv}{RGB}{0, 0, 0}

\sethlcolor{lightgray}

\makeatletter
\DeclareRobustCommand{\iscircle}{\mathord{\mathpalette\is@circle\relax}}
\newcommand\is@circle[2]{%
  \begingroup
  \sbox\z@{\raisebox{\depth}{$\m@th#1\bigcirc$}}%
  \sbox\tw@{$#1\square$}%
  \resizebox{!}{\ht\tw@}{\usebox{\z@}}%
  \endgroup
}
\makeatother

\IEEEoverridecommandlockouts                              % This command is only needed if 
\newcommand{\rom}[1]{\uppercase\expandafter{\romannumeral #1\relax}}
\title{\LARGE \bf TRIP: Terrain Traversability Mapping With Risk-Aware Prediction for Enhanced Online Quadrupedal Robot Navigation}

\author{
Minho Oh$^{1}$, Byeongho Yu$^{1}$, I Made Aswin Nahrendra$^{1}$, Seoyeon Jang$^{1}$, Hyeonwoo Lee$^{1}$, Dongkyu Lee$^{1}$, Seungjae Lee$^{1}$, Yeeun Kim$^{1}$, Marsim Kevin Christiansen$^{1}$, Hyungtae Lim$^{2}$, and Hyun Myung$^{1\ast}$
\thanks{$^*$Corresponding author: Prof. Hyun Myung}
\thanks{$^1$Urban Robotics Lab., School of Electrical Engineering, KAIST, Daejeon, 34141, Republic of Korea.  
{\tt\scriptsize \{minho.oh, bhyu, anahrendra, 9uantum01, hyeonwoolee, dklee, sj98lee, yeeunk, kevinmarsim, hmyung\}@kaist.ac.kr}
}
\thanks{$^2$Laboratory for Information and Decision Systems (LIDS)., MIT, Cambridge, MA, USA. 
{\tt\footnotesize shapelim@mit.edu}
}

\vspace{-5mm}
}

\begin{document}
\maketitle
% \thispagestyle{empty}
% \pagestyle{empty}
%%%%%%%%%%%%%%%%%%%%%%%%%%%%%%%%%%%%%%%%%%%%%%%%%%%%%%%%%%%%%%%%%%%%%%%%%%%%%%%%
% \markboth
% {IEEE Robotics and Automation Letters. Preprint Version. Month 2024}
% {Oh \MakeLowercase{\textit{et al.}}: TRIP: Terrain Traversability Mapping with Risk-Aware Predictions for Enhanced Online Quadrupedal Robot Navigation} 
\IEEEpeerreviewmaketitle
\begin{abstract}
    % \textcolor{red}{(WHY)}
    Accurate traversability estimation using an online dense terrain map is crucial for safe navigation in challenging environments like construction and disaster areas. 
    % % \textcolor{red}{(Problem)}
    However, traversability estimation for legged robots on rough terrains faces substantial challenges owing to limited terrain information caused by restricted field-of-view, and data occlusion and sparsity. 
    To robustly map traversable regions, we introduce \textit{terrain traversability mapping with risk-aware prediction}~(TRIP).
    % \textcolor{red}{(Contributions)}
    TRIP reconstructs the terrain maps while predicting multi-modal traversability risks, enhancing online autonomous navigation with the following contributions.
    Firstly, estimating steppability in a spherical projection space allows for addressing data sparsity while accomodating scalable terrain properties.
    Moreover, the proposed \textit{traversability-aware Bayesian generalized kernel}~(T-BGK)-based inference method enhances terrain completion accuracy and efficiency.
    Lastly, leveraging the steppability-based Mahalanobis distance contributes to robustness against outliers and dynamic elements, ultimately yielding a static terrain traversability map.
    % \textcolor{red}{(Eval)}
    As verified in both public and our in-house datasets, our TRIP shows significant performance increases in terms of terrain reconstruction and navigation map.
    % Our experiments comprehensively assess TRIP's performance in terrain reconstruction and navigation map perspectives, introducing the proposed ground-truth terrain traversability map. TRIP's results underscore its contributions, leveraging both public datasets and our own datasets. 
    A demo video that demonstrates its feasibility as an integral component within an onboard online autonomous navigation system for quadruped robots is available at \hyperlink{https://youtu.be/d7HlqAP4l0c}{https://youtu.be/d7HlqAP4l0c}.
    % Additionally, our approach demonstrates its feasibility as an integral component within an onboard online autonomous system for quadruped robots via demo video: \hyperlink{https://youtu.be/d7HlqAP4l0c}{https://youtu.be/d7HlqAP4l0c}.
\end{abstract}
% \vspace{-1mm}
%%%%%%%%%%%%%%%%%%%%%%%%%%%%%%%%%%%%%%%%%%%%%%%%%%%%%%%%%%%%%%%%%%%%%%%%%%%%%%%%
% \vspace{-3mm}
\section{Introduction}

    % \textcolor{red}{기존 연구내용들 및 장단점에 대해 개조식 작성}
    % Traversability mapping에 대한 도입
    \IEEEPARstart{T}{raversability} mapping is one of critical modules for autonomous and remote robot navigation~\cite{frey2022locomotion, grandia2023perceptive}.
    For legged robots navigating through irregular environments to perform a mission, such as construction sites and disaster areas,
    generating accurate and dense terrain maps in real-time becomes more essential~\cite{miki2022elevationmapcupy, xue2023traversability}.
    \begin{figure}[t!]
        \captionsetup{font=footnotesize}
        \centering
        \includegraphics[width=\columnwidth]{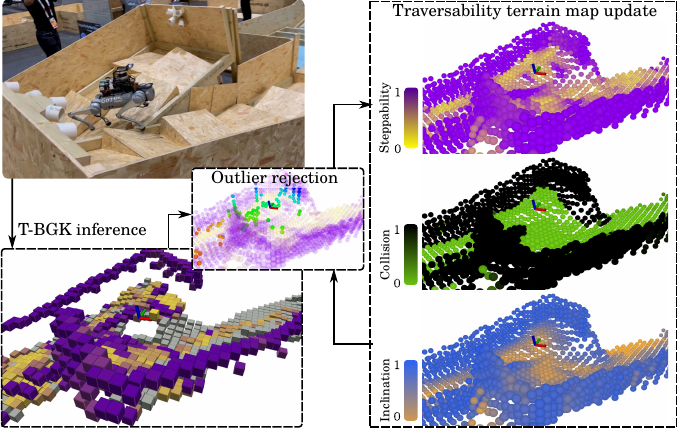}
        \caption{Overview of \textit{terrain traversability mapping with risk-aware prediction}~(TRIP). TRIP generates a local and global terrain map with multi-modal traversability risk prediction, enhancing online quadruped robot navigation. All figures in this paper are best viewed in color.}
        \label{fig:overview}
        % \vspace{-6mm}
    \end{figure}
    \begin{figure*}[t!]
        \captionsetup{font=footnotesize}
        \centering
        \includegraphics[width=\textwidth]{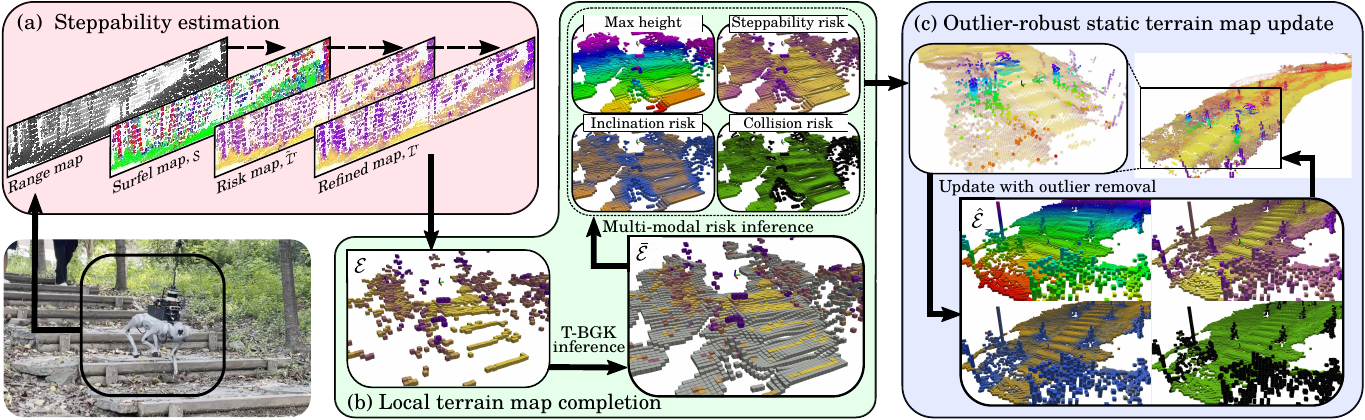}
        \caption{Overview of the proposed TRIP framework. 
                (a)~Using 3D LiDAR or depth camera, a surfel map $\mathcal{S}$ is generated from a range map. And the surfel map is used to build a steppability risk maps $\tilde{\mathcal{I}}^{r}$ and $\mathcal{I}^{r}$.
                (b)~After projecting the surfel and the steppability risk onto the elevation map $\mathcal{E}$, the local terrain map $\bar{\mathcal{E}}$ is completed based on traversability-aware Bayesian generalized kernel~(T-BGK) $k^{\mathcal{T}}$ and inference function $\mathcal{L}^{\mathcal{T}}$ while embedding traversability risks.
                (c)~Outliers are rejected using steppability-based Mahalanobis distance, and a static terrain map $\hat{\mathcal{E}}$ is updated with our proposed bias models.
                Each risk layer in the terrain map is color-coded to represent varying risk levels. 
                Yellow, orange, and green shades indicate low risk levels, while purple, blue, and black denote high-risk areas. 
                This color scheme is consistently employed for terrain map figures throughout this paper.
                % Each risk layer of terrain map is color-coded to denote risk levels; 
                % proximity to yellow, orange, and green signifies low risk, while purple, blue, and black indicates high risks.
                % This color sheme is commonly used for terrain map figures in this paper.
                }
        \label{fig:fullframework}
        \vspace{-5mm}
    \end{figure*}

    % terrain elevation map 연구의 필요성
    % 기존 navigation을 위한 map으로는 2D / 3D를 많이 사용
    % 최근 quadruped robot에 대한 연구가 늘어나면서, elevation map 기반의 지도를 다양한 분야에서 활용 중.
    Traditional navigation maps for robots have employ 2D occupancy grid maps for wheeled platforms and 3D voxel maps for flying ones. 
    However, recent research into quadruped robots, exploring underground or construction environments demands more detailed terrain representation~\cite{frankhauser2018elevationmap}.
    Such detailed terrain information can be directly leveraged in the navigation algorithms of legged robots, such as 
    trajectory optimization~\cite{michael2017trajectoryopt, alexander2018gaitandtrajectoryopt, matias2022efficient, yang2023iplanner, chen2023smugplanner}, 
    footstep planning~\cite{griffin2019footstep, bertrand2020usableplane, grandia2021multi}, 
    and locomotion control~\cite{jenelten2020perceptive, kim2020vision, miki2022learning, jenelten2022tamols}.
    
    % 하지만, 한계점이 존재함: insufficient terrain information, stemming from limitations in field-of-view, data occlusion, and data sparsity
    However, when ground platforms traverse rough terrains equipped with range sensors such as 3D LiDAR and depth cameras, challenges in terrain map arise owing to insufficient terrain information.
    This insufficiency is caused by limitations in the field-of-view, data occlusion, and data sparsity, leading to compromised locomotion capabilities~\cite{yang2023neuraldense, zhang2022vision, hoeller2022neuralscene, shan2018bayesian}.
    Moreover, for safer navigation, separate traversability estimation modules for to be embedded sequentially in the terrain map~\cite{griffin2019footstep, bertrand2020usableplane}, which leads to consume considerable processing time.
    These sequential approaches, which do not consider the traversability in terrain reconstruction, also sacrifices the terrain reconstruction accuracy.
    Additionally, a key challenge in off-road navigation is how to handle visually-similar terrains within the same semantic category, which may exhibit substantially different traversability properties~\cite{cai2022probabilistic, cai2022risk}.

    % 우리의 contributions
    To tackle these challenges, we focus on constructing a terrain traversability map that enables safe navigation for quadrupedal robots~(Fig.~\ref{fig:overview}).
    We propose TRIP, an online approach that enhances terrain reconstruction and navigation cost estimation performances simultaneously, even in visually similar terrains that may have different traversable properties owing to terrain-specific risks with the following contributions:
    \begin{itemize}
        % \vspace{-1mm}
        \item {
            Our steppability estimation, which estimate geometric properties in spherical projection space, enhances scalable terrain map completion in environments from narrow spaces to open areas, addressing the sparsity issues.
        }
        \item {
            Our proposed \textit{traversability-aware Bayesian generalized kernel}~(T-BGK)-based inference enhances local terrain map completion while emphasizing traversability risks, thereby assisting in the avoidance of hazardous terrains.
        }
        \item {
            Steppability-based Mahalanobis distance filter guarantees the robustness of the terrain map against outliers originating from dynamic objects and other sources of noise.
        }
    \end{itemize}

\newpage
\section{Related Work}
    \vspace{-1mm}
    \subsection{Terrain Reconstruction}
    \vspace{-1mm}
    Recent studies in terrain reconstruction for quadrupedal robots that rely on 2.5D elevation maps can be classified into the following three categories:
    \subsubsection{Filter-Based Approaches}
    Fankhauser~\textit{et al.}~\cite{frankhauser2018elevationmap} proposed a 2.5D elevation map with confidence bounds, updated using a 1D Kalman filter at the cell level, forming the basis for locomotion control.
    Zhang~\textit{et al.}~\cite{zhang2022vision} introduced a robust terrain map reconstruction algorithm resilient to localization drift, providing traversability information for each cell based on $xy$-plane slope.
    However, these methods face challenges arising from data sparsity, which may compromise subsequent navigation performance.
    
    \subsubsection{Probability-Based Approaches}
    To address data sparsity challenges, probability-based terrain inference approaches have been proposed. 
    Doherty~\textit{et al.}~\cite{kevin2017bayesian, kevin2019learning} introduced Bayesian generalized kernel~(BGK)-based 3D occupancy map prediction method for real-time inference of empty voxels based on neighboring voxels.
    To solve this sparsity issue, Shan~\textit{et al.}~\cite{shan2018bayesian} utilized BGK-based inference to create a 2.5D dense global terrain map with traversability cost, addressing sparse range sensor data.
    They use height information from adjacent cells to estimate traversability but do not utilize traversability information reciprocally.
    This oversight leads to inaccurate terrain reconstruction in areas with little observed data. Moreover, applying BGK-based inference on a global terrain map results in a processing time increase.
    % They use height inform
    % Therefore, we propose also leveraging traversability information in terrain reconstruction to improve inference performance.
    % Consequently, this approach leads to inaccurate inferences in areas with insufficient data.
    % However, when estimating the traversability of an empty cell, the height information of surrounding cells is used. But on the contrary, since traversability is not used to estimate terrain information, there is an area of inaccurate inference when data is insufficient.

    \subsubsection{Learning-Based Approaches}
    Learning-based methods have also emerged to enhance terrain mapping in challenging situations where sensors face limitations or are obstructed by the robot's systems.
    Hoeller~\textit{et al.}~\cite{hoeller2022neuralscene} proposed an encoder-decoder network for accurate terrain representation, especially with noisy depth camera setting.
    Yang~\textit{et al.}~\cite{yang2023neuraldense} employed a learned model to obtain binary edge and dense elevation maps with uncertainty, addressing sparsity in 3D LiDAR. 
    Yet, these methods have limited applications in structured environments used for learning and may encounter errors owing to dynamic obstacles or overhanging obstacles.
    
    \subsection{Traversability Estimation}
    % To enhance the navigation equipped with range sensors, research on traversability estimation often complements elevation mapping efforts~\cite{miki2022elevationmapcupy} and includes:
    To enhance the navigation equipped with range sensors, some studies involve traversability estimation on the elevation mapping through the following approaches:
    % \subsubsection{Incorporating Terrain Data}
    % \subsubsection{Incorporating Geometric Properties of Terrain}
    \subsubsection{Statistical Approaches}
    Wermelinger~\textit{et al.}~\cite{fankhauser2014robot, wermelinger2016navigationplanning} proposed a navigation algorithm leveraging a traversability map calculated using local slope, roughness, and step height.
    Kim~\textit{et al.}~\cite{kim2020vision} incorporated steppability into footstep planning, estimated based on slopes.
    Xue~\textit{et al.}~\cite{xue2023traversability} assessed traversability by deriving local convexity from elevation and normal maps for autonomous vehicles.
    
    \subsubsection{Self-supervised Approaches}
    To achieve robust traversability estimation in unknown off-road environments, some researchers have introduced self-supervised approaches with elevation maps.
    Cai~\textit{et al.}~\cite{cai2022probabilistic, cai2022risk} suggested terrain traversability analysis using a self-supervised learning model to learn empirical traction parameter distributions.
    Frey~\textit{et al.}~\cite{frey2023fast} introduced a vision-based self-supervised learning method to enhance traversability estimation in the wild, which is integrated into the elevation map.
    
    Traversability estimation methods vary based on the specific platform and environmental factors in consideration, as above. In our work, instead of solely defining traversability, we propose a complementary approach that integrates terrain map reconstruction and traversability estimation. This approach is designed to cater to a broader range of environments, emphasizing versatility in its application.

%%%%%%%%%%%%%%%%%%%%%%%%%%%%%%%%%%%%%%%%%%%%%%%%%%%%%%%%%%%%%%%%%%%%%%%%%%%%%%%%
% \vspace{-3mm}
\section{Terrain Traversability Mapping}
    As depicted in Fig.~\ref{fig:fullframework}, TRIP consists of three primary steps for predicting a terrain map with traversability:
    a)~Steppability estimation from range data in spherical projection space,
    b)~local terrain map completion using traversability-aware terrain inference method, and
    c)~terrain map update with steppability-based Mahalanobis distance-based outlier rejection.
    Following these procedures, we improve the terrain map to be resilient to spatial scale changes and outliers, featuring three distinct traversability risks. 
    This enhanced terrain map proves valuable for the online navigation system of quadruped robots.

\newpage
\subsection{Steppability Estimation in Spherical Projection Space}
    \begin{figure}[b!]
        \vspace{-2mm}
        \captionsetup{font=footnotesize}
        \centering
        \includegraphics[width=\columnwidth]{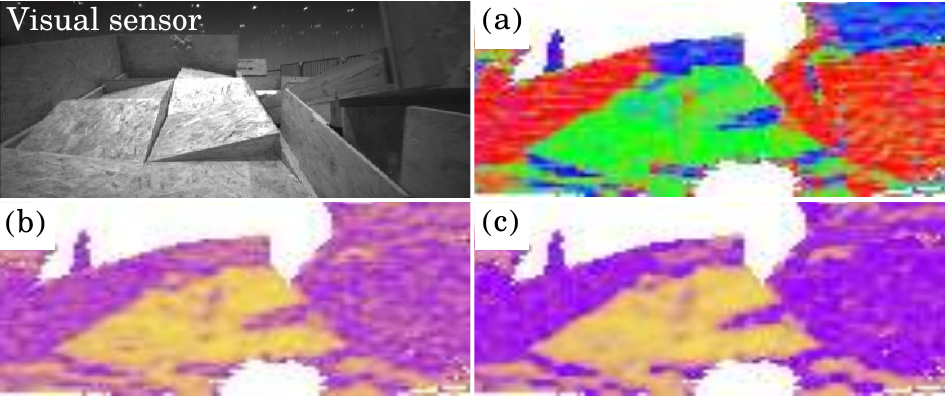}
        \caption{Steppability map results from (a)~surfel map $\mathcal{S}$ from the range sensor with $1\text{~pixel}=1^\circ \times 1^\circ$ resolution: 
        (b)~Raw risk map $\tilde{\mathcal{I}}^{r}$, refined risk maps with (c)~our conditional pooling $\mathcal{I}^{r}$.
        Our conditional pooling successfully reduces noise and enhances the risk discernment.
        }
        \label{fig:spherical_space_pooling}
    \end{figure}
    In this section, we employ an efficient coarse-to-fine approach for steppability estimation from range data in spherical projection space~\cite{oh2022travel}, as illustrated in the sequence shown in Fig.~\ref{fig:fullframework}(a).
    Leveraging the spherical projection space is advantageous for handling sparse 3D data and allows for scalable terrain characteristics regardless of narrow or wide environments, 
    as opposed to grid-based approaches that vary in computational load depending on the distribution of the point cloud in the surroundings~\cite{grandia2023perceptive}.
    Building upon the approach in~\cite{behley2018suma}, we address geometrical properties by projecting the 3D data $(x,y,z)$ onto the pixel $(u,v)$ of the surfel map $\mathcal{S}\in\mathbb{R}^{w\cross h}$ as follows:
    \begin{equation}
        \begin{pmatrix}
        u \\ v
        \end{pmatrix} =
        \begin{pmatrix}            
            w\cdot[1-\frac{\arctan(y/x)+f_r}{f_l+f_r}]            
            \\            
            h\cdot[1-\frac{\arcsin(z/\sqrt{x^2+y^2})+f_b}{f_t+f_b}]
        \end{pmatrix},
    \end{equation}
    where $f_t + f_b$ corresponds to the vertical field-of-view, while $f_r + f_l$ represents the horizontal field-of-view.
    Each surfel $\mathbf{s}=\mathcal{S}(u,v)$ consists of point vector $\mathbf{p}=(p_x,p_y,p_z)^T\in\mathbb{R}^3$ and normal vector $\mathbf{n}=(n_x, n_y, n_z)^T\in\mathbb{R}^3$, where $\mathbf{n}$ is estimated by principal component analysis~(PCA) with the surrounding points within the corresponded kernel $\mathcal{K}^{\mathcal{I}}$~\cite{lim2021patchwork}.
    
    By taking $\mathcal{S}$, we define the steppability risk map $\tilde{\mathcal{I}}^{r}\in\mathbb{R}^{w\cross h}$, where each risk $\tilde{r}^{\mathrm{step}}=\tilde{\mathcal{I}}^{r}(u,v)$ is obtained through the geometric mean of verticality and proximity as follows:
    \begin{equation}
        \tilde{r}^{\mathrm{step}}
        = 1-\sqrt{\frac{n_{z}}{\mathrm{valid}\left(\mathcal{K}^{\mathcal{I}}\right)}{\sum_{(i,j)}^{\mathcal{K}^{\mathcal{I}}}{\mathrm{prox} \left(\mathbf{s}^{(i,j)},\mathbf{s}\right)}}},
    \end{equation}
    where $\mathrm{valid}(\cdot)$ counts the number of valid elements. Note that smaller $\tilde{r}^{\mathrm{step}}$ indicates a higher likelihood of a robot being steppable. 
    Our proposed proximity function, $\mathrm{prox}(\cdot,\cdot)$, is introduced to quantitatively measure both the distance and convexity ratio between surfels as follows by modifying $lcc(\cdot)$, which was proposed in our previous work~\cite{oh2022travel}:
    \begin{multline}
        \mathrm{prox}\left(\mathbf{s}^{\alpha},\mathbf{s}^{\beta}\right) \in[0,1] 
            \\
            = \left|\mathbf{n}^{\alpha}\cdot\mathbf{n}^{\beta}\right|
              \frac
              {\max\left(\left|\mathbf{n}^{\alpha} \cdot (\mathbf{p}^{\beta} - \mathbf{p}^{\alpha})\right|, \left|\mathbf{n}^{\beta} \cdot (\mathbf{p}^{\alpha} - \mathbf{p}^{\beta}) \right|\right)}
              {\left|\mathbf{p}^{\beta}-\mathbf{p}^{\alpha}\right|}.
    \end{multline}
    However, $\mathcal{S}$ and $\tilde{r}^{\mathrm{step}}$ are susceptible to sensor noise, as shown in Figs.~\ref{fig:spherical_space_pooling}(a)~and~~\ref{fig:spherical_space_pooling}(b), leading to ambiguity in distinction. To address this issue and enhance our subsequent procedures, we propose a conditional pooling method designed to reduce errors on the terrain while simultaneously preserving the risk edges as follows:
    \begin{align}
        r^{\mathrm{step}} = 
            \begin{cases} 
            \max\left(\{\tilde{r}^{\mathrm{step}}_{(i,j)\in\mathcal{K}^{\mathcal{I}}}\}\right)  &,\mathrm{if}~\mu^{\mathrm{step}} > \tau^{r} 
            \\                            
            \mu^{\mathrm{step}} &,\mathrm{otherwise}                            
            \end{cases},
    \end{align}
    where $\mu^{\mathrm{step}}$ and $\tau^{r}$ are the mean value of $\{\tilde{r}^{\mathrm{step}}_{(m,n)\in\mathcal{K}^{\mathcal{I}}}\}$ and a threshold value, respectively.
    The steppability risk map $\mathcal{I}^{r}=\{r^{\mathrm{step}}\in[0,1]\}$ is shown in Fig.~\ref{fig:spherical_space_pooling}(c).
    % where $\mu^{\mathrm{step}}=\mathrm{E}\left(\{\tilde{r}^{\mathrm{step}}_{(m,n)\in\mathcal{K}^{\mathcal{I}}}\}\right)$ and $\tau^{r}$ is a threshold. The steppability risk map $\mathcal{I}^{r}=\{r^{\mathrm{step}}\in[0,1]\}$ is shown in Fig.~\ref{fig:spherical_space_pooling}(c).

\vspace{-2mm}
\subsection{Local Terrain Map Completion Based on T-BGK}
\vspace{-1mm}

    % Akin to the procedure illustrated between Figs.~\ref{fig:fullframework}(a) and~\ref{fig:fullframework}(b), 
    Next, $\mathcal{S}$ and $\mathcal{I}^{r}$ are re-projected onto the terrain map 
    $\mathcal{E}=\{(\mathbf{o}_{\mathbf{e}}, h_{\mathbf{e}}^{\max}, h_{\mathbf{e}}^{\min}, n^{z}_{\mathbf{e}}, r^{\mathrm{step}}_{\mathbf{e}})\}$. 
    Here, $\mathbf{e}$ and $o_{\mathbf{e}}$ denote the index and the central location in $xy$-coordinate for each cell of terrain map, respectively.
    
    Note that multiple elements from the terrain and overhanging objects may correspond to the same $\mathbf{e}$ simultaneously. To reject overhanging elements, which can compromise the accuracy of the terrain map~\cite{miki2022elevationmapcupy}, re-projection is executed from the bottom of $\mathcal{S}$ and $\mathcal{I}^{r}$ upwards. During this process, elements are filtered out if their $p_z$ differs by more than the platform height~$h_p$ from the height of the corresponding cell~$h_{\mathbf{e}}^{\max}$.
    
\begin{figure}[t!]
        \vspace{-2mm}
        \captionsetup{font=footnotesize}
        \centering
        \includegraphics[width=\columnwidth]{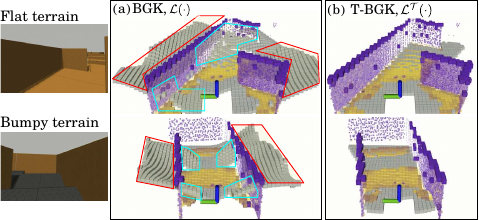}
        \caption{
                Local inference results, $\bar{\mathcal{E}}$, on the enclosed flat and bumpy terrains: 
                (a)~Vanila BGK-based inference $\mathcal{L}(\cdot)$ results. (b)~The proposed T-BGK-based inference $\mathcal{L}^{\mathcal{T}}(\cdot)$ results.
                The grey points represent the predicted results of each inference function.
                Vanila BGK infers unobservable regions beyond walls~(red areas) and uses every neighbors without any consideration of steppability when predicting the empty cells~(cyan areas). 
                In contrast, the proposed T-BGK distinguishes observable regions and is based on $r^{\mathrm{step}}$, allowing for traversability-aware predictions.
                }
        \label{fig:tbgk}
        \vspace{-5mm}
    \end{figure}

    However, as shown in Fig.~\ref{fig:fullframework}(b), the naively re-projected terrain map~$\mathcal{E}$ appears sparse and discrete with noticeable gaps, particularly during stair descent, which is insufficient for visual locomotion or navigation.
    To handle the sparsity issue, Shan~\textit{et al.}~\cite{shan2018bayesian} applied an approach based on BGK~$k(\cdot,\cdot)$ and its inference function~$\mathcal{L}(\cdot)$, as proposed in \cite{melkumyan2009sparse} as follows:
    \begin{multline}
        k\left(\mathbf{e}^{\alpha},\mathbf{e}^{\beta}\right)=
        \\
        \begin{cases}
        \frac{2+\cos(2\pi d/l)}{3}(1-d/l)+\frac{\sin(2\pi d/l)}{2\pi}, &\mathrm{if}~d/l < 1
        \\
        0 , &\mathrm{otherwise}
        \end{cases},
        \label{eq:bgk}
    \end{multline}
    \begin{equation}
        \mathcal{L}\left(\mathcal{K}^{\mathcal{E}}_{\mathbf{e}}\right)
        \triangleq \bar{y}_{\mathbf{e}} 
        = \frac{\sum_{\mathbf{e}^{i}}^{\mathcal{K}^{\mathcal{E}}_k} k(\mathbf{e}^{i},\mathbf{e})\cdot y_{\mathbf{e}^{i}}}{\sum_{\mathbf{e}^{i}}^{\mathcal{K}^{\mathcal{E}}_k} k(\mathbf{e}^{i},\mathbf{e})}, %, y \in \{r^{\mathrm{step}}_{i}\}
        \label{eq:bgk_inference}
    \end{equation}
    where $d=||\mathbf{o}_{\mathbf{e}^\alpha}-\mathbf{o}_{\mathbf{e}^\beta}||$ and $l$ is the radius of the prediction kernel $\mathcal{K}^{\mathcal{E}}_k$. 
    % $\mathbf{o}$ denotes the $xy$-location of the center of $\mathbf{e}$ index. 
    As shown in Fig.~\ref{fig:tbgk}(b), BGK utilizes neighbor cells without any consideration of steppability, resulting in imprecise terrain mapping for cells near the walls. 
    % As shown in Fig.~\ref{fig:tbgk}(b), BGK utilizes neighbor cells indiscriminately, resulting in incorrect inferences for cells near the walls. 
    Additionally, it infers unobservable regions beyond walls, potentially affecting the updating of the terrain map. 
    
    To that end, we propose \textit{T-BGK} $k^{\mathcal{T}}(\cdot, \cdot)$ and inference function $\mathcal{L}^{\mathcal{T}}(\cdot)$ by modifying (\ref{eq:bgk})~and~(\ref{eq:bgk_inference}) to account for steppability as follows:
    \begin{equation}
        k^{\mathcal{T}}\left(\mathbf{e}^{\alpha},\mathbf{e}^{\beta}\right)=
            (1-r^{\mathrm{step}}_{\mathbf{e}^{\beta}})k(\mathbf{e}^{\alpha},\mathbf{e}^{\beta}),
        \label{eq:tbgk}
    \end{equation}
    \begin{equation}
        \mathcal{L}^{\mathcal{T}}\left(\mathcal{K}^{\mathcal{E}}_{\mathbf{e}}\right) \triangleq
        \bar{y}_{\mathbf{e}}
        = \frac{\sum_{\mathbf{e}^{i}}^{\mathcal{K}^{\mathcal{E}}_\mathbf{e}} k^{\mathcal{T}}(\mathbf{e}^{i},\mathbf{e})y_{\mathbf{e}^{i}}}{\sum_{\mathbf{e}^{i}}^{\mathcal{K}^{\mathcal{E}}_\mathbf{e}} k^{\mathcal{T}}(\mathbf{e}^{i},\mathbf{e})}.
        \label{eq:tbgk_inference}
    \end{equation}

    T-BGK-based completion method infers $h^{\max}$ and $h^{\min}$, while the steppability risk map $r^{\mathrm{step}}$ is inferred by vanilla BGK, similar to Shan~\textit{et al.}~\cite{shan2018bayesian}.
    Furthermore, the inference region of T-BGK is bounded by maximum ranges within each column of $\mathcal{S}$.
    Consequently, as shown in Fig.~\ref{fig:tbgk}(c), these features allow for precise terrain prediction within observable regions, improving spatial reconstruction accuracy.
    
    For the inferred cells, the verticality $\hat{n}^{z}$ is simply calculated by applying PCA to the set of location and their maximum height for each cell in the corresponding kernel $\{{\mathbf{o}^{i}, \bar{h}^{\max}_{\mathbf{e}^i}}\}_{\mathbf{e}^{i}\in\mathcal{K}^{\mathcal{E}}_{\mathbf{e}^{i}}}$. 
    Additionally, the inclination risk $\bar{r}^{\mathrm{incl}}$ and collision risk $\bar{r}^{\mathrm{coll}}$, which indicate the navigational costs of each area, are embedded on the dense terrain map as follows:
    \begin{align}
        \bar{r}^{\mathrm{incl}}_{\mathbf{e}} 
        &= \max\left(
        \{\arcsin(\frac{\left|\bar{h}^{\max}_{\mathbf{e}} - \bar{h}^{\max}_{\mathbf{e}^{i}}\right|}
                        {\lVert \mathbf{o}_{\mathbf{e}}-\mathbf{o}_{\mathbf{e}^{i}}\rVert_2})
        \}_{{\mathbf{e}^{i}}\in \mathcal{K}^{\mathcal{E}}_{\mathbf{e}}}
        \right)/2\pi,
        \\
        \bar{r}^{\mathrm{coll}}_{\mathbf{e}} &= 
        \min\left(\max\left(\{\bar{h}^{\max}_{\mathbf{e}^{i}}-\bar{h}^{\min}_{\mathbf{e}^{i}}\}_{\mathbf{e}^{i}\in\mathcal{K}^{\mathcal{E}}_{\mathbf{e}}}\right)/\tau^{h},1\right),
    \end{align}
    where $\bar{h}^{\max}_{\mathbf{e}}$, $\bar{h}^{\min}_{\mathbf{e}}$, and $\bar{r}^{\mathrm{step}}_{\mathbf{e}}$ are the results of T-BGK for each cell. 
    As a result of our proposed local terrain map completion, 
    the inferred local terrain $\bar{\mathcal{E}}$ is comprised of $\{\bar{h}^{\max}_{\mathbf{e}}, \bar{r}^{\mathrm{step}}_{\mathbf{e}}, \bar{r}^{\mathrm{incl}}_{\mathbf{e}}, \bar{r}^{\mathrm{coll}}_{\mathbf{e}}\}$ layers, as illustrated in the dashed box of Fig.~\ref{fig:fullframework}(b).
    In order to define the uncertainties of our inference function, we propose two bias models of T-BGK:
    A horizontal bias $\sigma^{\mathbf{o}}_{i}$, which is a sum of horizontal signed $xy$-vectors, and a vertical bias $\sigma^{h}_{i}$, which is a sum of $h^{\max}$ differences as follows:
    \begin{align}
        \sigma^{\mathbf{o}}_{\mathbf{e}} &= 
        \left|
            \frac
            {\sum_{\mathbf{e}^{i}}^{\mathcal{K}^{\mathcal{E}}_{\mathbf{e}}} k^{\mathcal{T}}(\mathbf{e}^{i},\mathbf{e})(\mathbf{o}_{\mathbf{e}^{i}}-\mathbf{o}_{\mathbf{e}})}
            {\sum_{\mathbf{e}^{i}}^{\mathcal{K}^{\mathcal{E}}_{\mathbf{e}}} k^{\mathcal{T}}(\mathbf{e}^{i},\mathbf{e}))}
        \right|
        \in[0,1],
        \label{eq:horizon_bias}
        \\
        \sigma^{h}_{\mathbf{e}} &= 
        \frac
        {\sum_{\mathbf{e}^{i}}^{\mathcal{K}^{\mathcal{E}}_{\mathbf{e}}} k^{\mathcal{T}}(\mathbf{e}^{i},\mathbf{e})\left|h^{\max}_{\mathbf{e}^{i}}-\bar{h}^{\max}_{\mathbf{e}}\right|}
        {\sum_{\mathbf{e}^{i}}^{\mathcal{K}^{\mathcal{E}}_{\mathbf{e}}} k^{\mathcal{T}}(\mathbf{e}^{i},\mathbf{e})}
        \in[0,1].
        \label{eq:vertical_bias}
    \end{align}
    % As a result, taking the advantage to relax the independent cell assumption, our T-BGK method brings smoothness and continuity to the terrain map completion with risk information.
    % \begin{figure}[b!]
    %     \vspace{-3mm}
    %     \captionsetup{font=footnotesize}
    %     \centering
    %     \includegraphics[width=\columnwidth]{figures/TRIP_figure5_embedding_vf.pdf}
    %     \caption{Risk layers of the inferred local terrain map $\bar{\mathcal{E}}$. 
    %             (L-R): Maximum height $h^{\max}$, steppability risk $r^{\mathrm{step}}$, inclination risk $r^{\mathrm{incl}}$, and collision risk $r^{\mathrm{coll}}$.}
    %     \label{fig:embed}
    % \end{figure}
    These bias models are leveraged to enhance the terrain map updates as the measurement noise models, which are differentiated according to reliability of each elements.
    Therefore, the completed terrain map is generated as follows:
    \begin{equation}
        \begin{split}
            \bar{\mathcal{E}} = 
            \{(\mathbf{o}_{\mathbf{e}}, \bar{h}^{\max}_{\mathbf{e}}, \bar{h}^{\min}_{\mathbf{e}}, \bar{n}^{z}_{\mathbf{e}}, \bar{r}^{\mathrm{step}}_{\mathbf{e}}, \bar{r}^{\mathrm{incl}}_{\mathbf{e}}, \bar{r}^{\mathrm{coll}}_{\mathbf{e}})\} \\
            \text{with~} \{\sigma^{\mathbf{o}}_{\mathbf{e}},\sigma^{h}_{\mathbf{e}}\}.
        \end{split}
    \end{equation}
% \newpage
\subsection{Outlier-Robust Static Terrain Map Update}~\label{sec:map_update}
% \vspace{-4mm}
    \begin{figure}[t!]
        \vspace{-2mm}
        \captionsetup{font=footnotesize}
        \centering
        \includegraphics[width=\columnwidth]{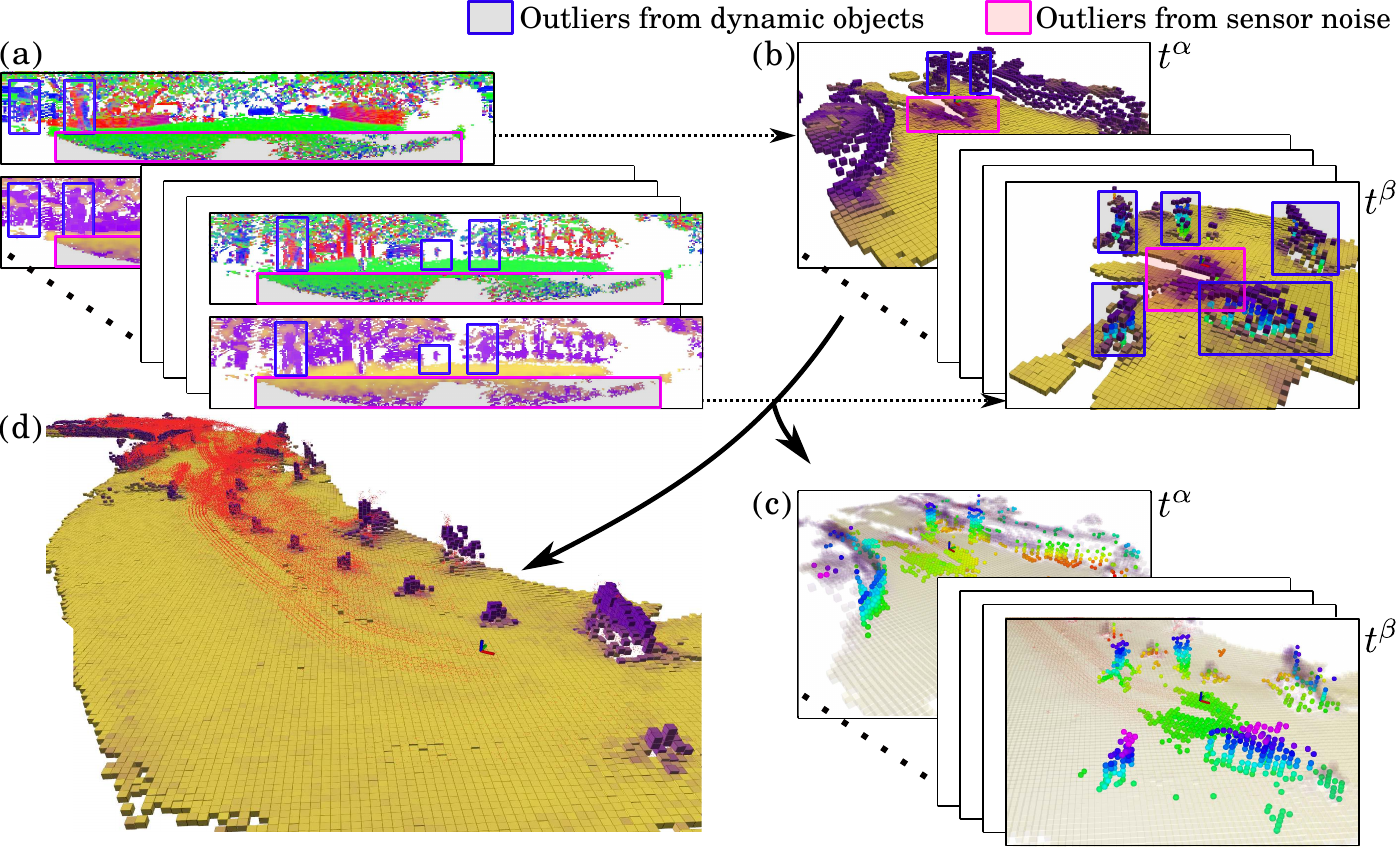}
        \caption{
                Example of outliers and sequences of our steppability-based Mahalanobis distance filtering in a real-world environment.
                (a)~The surfel map and the steppability risk map.
                Among (b)~the local terrain maps which are completed for each time, (c)~the outliers are filtered based on the updated terrain map at the previous time.
                (d)~After the rejection, the terrain map is updated based on Kalman filter.
                The red points represent the traces of rejected outliers.
                Our proposed rejection-based map update module enhances robustness of our terrain map against dynamic elements and sensor noises.
                }
        % \caption{
        %         Example of outliers and sequences of our steppability-based Mahalanobis distance filtering in a real-world environment.
        %         (a)~The surfel map $\mathcal{S}_{t^\alpha}$ and the steppability risk map $\mathcal{I}^{r}_{t^\alpha}$ at time ${t^\alpha}$.
        %         Among (b)~the local terrain map $\bar{\mathcal{E}}_{t^\alpha}$ and $\bar{\mathcal{E}}_{t^\beta}$, (c)~the outliers are filtered based on the previous updated terrain map $\hat{\mathcal{E}}_{t^{\alpha-1}}$ and $\hat{\mathcal{E}}_{t^{\beta-1}}$.
        %         (d)~After rejection, the terrain map $\hat{\mathcal{E}}_{t^{\beta}}$ is updated based on Kalman filter.
        %         The red points represent the traces of rejected outliers.
        %         Our proposed rejection-based map update module enhances robustness of our terrain map against dynamic elements and sensor noises.
        %         }
        \label{fig:outlier_filter}
        \vspace{-5mm}
    \end{figure}
    Referring the local completed terrain map $\bar{\mathcal{E}}$ in Fig.~\ref{fig:fullframework}(b), there exists an unobservable yet essential terrain area, notably underneath the platform. 
    Moreover, outliers are often encountered in real-world owing to inaccuracies induced by sensor noise, dynamic objects, and flying points from the range sensor~\cite{bertrand2020usableplane}, as shown in Figs.~\ref{fig:outlier_filter}(a)-(c). 
    These challenges affect the navigation performance; thus, it is necessary to reduce these undesirable effects on the local map.
    % These challenges faced by the local map disadvantageously impact the navigation performance.

    To this end, the outlier-robust map update module is proposed so that reflects temporal information and to output a static terrain traversability map, as shown in Fig.~\ref{fig:fullframework}(c).
    % To prevent subsequent navigation from encountering such problems, the outlier-robust map update module is proposed to reflect temporal information and to make the static terrain traversability map, as shown in Fig.~\ref{fig:fullframework}(c).
    Inspired by Song~\textit{et al.}~\cite{song2022dynavins}, Mahalanobis distance-based rejection method is proposed based on our steppability risk and verticality as follows:
    \begin{align}
        \hat{\mathbf{e}}(t)=
        &\begin{cases} 
            \mathrm{KF}(\hat{\mathbf{e}}(t-1), \mathbf{e}(t)) &,\:\:\mathrm{if}~d^{\mathcal{M}}_{\mathbf{e}(t)} < \tau^{\mathcal{M}}
            \\                            
            \hat{\mathbf{e}}(t)=\hat{\mathbf{e}}(t-1) &,\:\:\mathrm{otherwise}
        \end{cases},
        \label{eq:mahalanobis}
        \\
        d^{\mathcal{M}}_{\mathbf{e}(t)} =
            &   \begin{pmatrix}
                    \bar{n}^{z}_{\mathbf{e}(t)}     - \hat{n}^{z}_{\mathbf{e}(t-1)}   \\ 
                    \bar{r}^{\mathrm{step}}_{\mathbf{e}(t)}  -\hat{r}^{\mathrm{step}}_{\mathbf{e}(t-1)}
                \end{pmatrix}^{T}
                \Sigma^{-1}_{\mathbf{e}(t-1)}
                \begin{pmatrix}
                    \bar{n}^{z}_{\mathbf{e}(t)}     - \hat{n}^{z}_{\mathbf{e}(t-1)}   \\ 
                    \bar{r}^{\mathrm{step}}_{\mathbf{e}(t)}  -\hat{r}^{\mathrm{step}}_{\mathbf{e}(t-1)}
                \end{pmatrix},
    \end{align}
    where $\mathrm{KF}(\cdot,\cdot)$ and $d^{\mathcal{M}}_{\mathbf{e}(t)}$ are Kalman filter update and the Mahalanobis distance, respectively.
    The covariance matrix $\Sigma$ in (\ref{eq:mahalanobis}) is derived from Kalman filter update, which is used afterwards as map update, where $\mathbf{e}(t)$ denotes the index of the corresponding cell in $\bar{\mathcal{E}}_t$.

    By taking the current outlier-rejected terrain map $\bar{\mathcal{E}}_{t}$, the each cell of terrain map is recursively updated over time $t$ by Kalman filter, utilizing the proposed biases $\sigma^{\mathbf{o}}_{\mathbf{e}(t)}$ and $\sigma^{h}_{\mathbf{e}(t)}$ which are presented in (\ref{eq:horizon_bias}) and (\ref{eq:vertical_bias}).
    In detail, $\bar{h}^{\max}_{\mathbf{e}(t)}$ and $\bar{h}^{\min}_{\mathbf{e}(t)}$ are updated with $\sigma^{h}_{\mathbf{e}(t)}$, and other terrain properties $\{\bar{n}^{z}_{\mathbf{e}(t)}, \bar{r}^{\mathrm{step}}_{\mathbf{e}(t)}, \bar{r}^{\mathrm{incl}}_{\mathbf{e}(t)}\}$ are updated with $\sigma^{\mathbf{o}}_{\mathbf{e}(t)}$ as the measurement noise models.
    Only for the collision risk $\bar{r}^{\mathrm{coll}}_{\mathbf{e}(t)}$, we utilize the logit-based update, which is employed in Shan~\textit{et al.}~\cite{shan2018bayesian}, as follows:
    \begin{equation}
        \hat{r}^{\mathrm{coll}}_{\mathbf{e}(t)} 
        = 1 - 1/ \left(1 + \exp{\sum_{m\in[0,t]}{\log{\frac{\bar{r}^{\mathrm{coll}}_{\mathbf{e}(m)}}{1-\bar{r}^{\mathrm{coll}}_{\mathbf{e}(m)}}}}} \right),
    \end{equation}
    such that $\hat{r}^{\mathrm{coll}}$ are bounded by $0$~(free) and $1$~(collision).
    Finally, the updated static terrain map at time $t$ can be represented as 
    \begin{equation}
        \hat{\mathcal{E}}_{t}=\{(\mathbf{o}_{\mathbf{e}(t)}\hat{h}^{\max}_{\mathbf{e}(t)}, \hat{h}^{\min}_{\mathbf{e}(t)}, \hat{n}^{z}_{\mathbf{e}(t)}, \hat{r}^{\mathrm{step}}_{\mathbf{e}(t)}, \hat{r}^{\mathrm{incl}}_{\mathbf{e}(t)}, \hat{r}^{\mathrm{coll}}_{\mathbf{e}(t)})\}.
    \end{equation}
    
%%%%%%%%%%%%%%%%%%%%%%%%%%%%%%%%%%%%%%%%%%%%%%%%%%%%%%%%%%%%%%%%%%%%%%%%%%%%%%%%
% \newpage
\section{Experiments}
    % \vspace{-1mm}
    To evaluate our terrain traversability mapping algorithm, we utilize the public urban dataset, and our own quadruped robot datasets, which encompass both simulation and real-world data.
    For our quadruped robot datasets, we employ a Unitree Go1 robot with DreamWaQ~\cite{aswin2023dreamwaq} as the locomotion controller and LVI-Q~\cite{marsim2023ral} as the odometry algorithm.
    Furthermore, we have proposed and applied evaluation metrics to facilitate quantitative comparisons in terrain map reconstruction and navigation map performance.
    
    \vspace{-4mm}
    \subsection{Dataset}
    \subsubsection{Quadruped Robot Challenge~(QRC) Environments}
    The QRC environments, as provided by~\cite{jacoff2023taking}, feature challenging and scalable sections, including hurdles with deformable sponge, slopes with irregular terrains, stairs, and so forth. 
    We evaluated the performance of our terrain reconstruction and traversability assessment on these irregular and challenging terrains.
    The data were collected using an Ouster OS0-128 3D LiDAR and Xsens MTI-30 IMU along the one-way cyan path shown in Fig.~\ref{fig:exp_gt_map}(a). 
    And the QRC dataset comprises QRC simulations and QRC London sequences from simulation and real-world competition environments, respectively.

    \subsubsection{SemanticKITTI Dataset}
    In order to prove TRIP's robustness against outliers and versatility in urban structured scenes, we leverage the SemanticKITTI dataset~\cite{behley2019semantickitti}, which includes the dynamic objects and outlier elements with ground-truth labels in various urban scenes. 
    Inspired by Lim~\textit{et al.}~\cite{lim2021erasor} that used partial sequences with the most dynamic points to efficiently evaluate the robustness against dynamic objects, we also use these partial sequences.
    % use only partial scenes rather than full sequences from SemanticKITTI dataset, which includes dynamic objects. 
    
    \subsubsection{Campus Dataset}
    Unlike the artificial and structured terrain environment mentioned above, our campus dataset covers various scenes ranging from unstructured rough terrain scenes, such as irregular stairs and protruding obstacles encountered in the forest, to structured urban scenes that include various dynamic objects, such as pedestrian, bicycle, and vehicles.
    This dataset was introduced to quantitatively evaluate performance in challenging real-world environments that may be encountered during actual robot navigation, collected using Ouster OS1-32 3D LiDAR and Xsens MTI-300 IMU.
    
    \begin{figure}[t!]
        \captionsetup{font=footnotesize}
        \centering
        \includegraphics[width=\columnwidth]{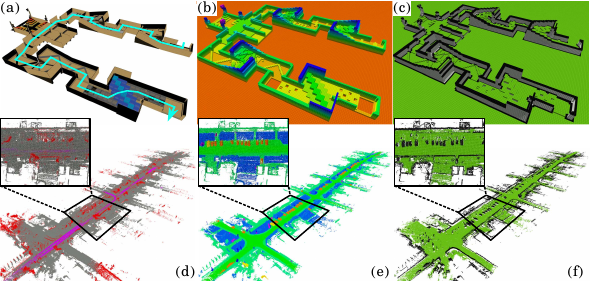}
        \caption{
                Ground-truth terrain traversability maps $\mathcal{E}^{gt}$ for quantitative comparison.
                (a)~The mesh model, (b)~its point cloud map, and ground truth map $\mathcal{E}^{gt}$ in the QRC environment~\cite{jacoff2023taking}.
                (d)~The naively accumulated point cloud map where purple points indicate \texttt{moving} points, red points indicate \texttt{unlabeled} or \texttt{outlier} points, and gray points indicate static points. 
                (e) The static point map and (f) the ground-truth map $\mathcal{E}^{gt}$ in the SemanticKITTI dataset~\cite{behley2019semantickitti}.
                % Regarding SemanticKITTI dataset~\cite{behley2019semantickitti}, (d)~by removing \texttt{unlabeled}, \texttt{outlier} and \texttt{moving} labeled points, (e)~the ground-truth static map was driven. Subsequently, $\mathcal{E}^{gt}$ for the real-world environment was semantically defined.
                % The upper maps depict simulation environment provided by QRC~\cite{jacoff2023taking}, while the lower maps show the \texttt{05}th sequence of SemanticKITTI dataset~\cite{behley2019semantickitti}.
                % Starting from the mesh model or semantic labeled map in the left column, the ground-truth static cloud map can be generated, as shown in the middle column. 
                % The right column maps represent the ground-truth 2.5D traversable terrain maps $\mathcal{E}^{gt}$, where green and black cells represent traversable and collision cells, respectively.
                }
        \label{fig:exp_gt_map}
        \vspace{4mm}
        \includegraphics[width=\columnwidth]{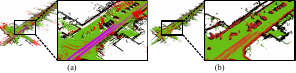}
        \caption{
                Terrain traversability map in the SemanticKITTI dataset~\cite{behley2019semantickitti}.
                (a)~The ground-truth traversability map with ground-truth dynamic elements in purple and outlier elements in red.
                (b)~The mapping result of our TRIP with the rejected outliers, which are represented in red points.
                }
        \vspace{-6mm}
        \label{fig:exp_robust}
    \end{figure}

    \vspace{-2mm}
    \subsection{Evaluation Approach}
    \vspace{-2mm}
    \subsubsection{Ground-truth 2.5D Terrain Traversability Map, $\mathcal{E}^{gt}$}
    For the quantitative assessment of terrain reconstruction and traversability performance, 
    we define $\mathcal{E}^{gt}$ as follows. %using data from QRC simulation and SemanticKITTI dataset.
    % we define $\mathcal{E}^{gt}$ using data from QRC simulation and SemanticKITTI dataset.

    In the QRC simulation, as depicted in Figs.~\ref{fig:exp_gt_map}(a)-(c), the 3D mesh model of the simulated terrain~(Fig.~\ref{fig:exp_gt_map}(a)) enables the generation of $\mathcal{E}^{gt}$~(Fig.~\ref{fig:exp_gt_map}(c)) based on the maximum height map~(Fig.~\ref{fig:exp_gt_map}(b)). Each cell is classified as a collision cell if the maximum height difference between the target cell and its adjacent cell exceeds $\tau^{h}$.

    For the SemanticKITTI dataset, ground-truth semantic labels are provided, facilitating to constructing $\mathcal{E}^{gt}$~(Fig.~\ref{fig:exp_gt_map}(f)). Particularly, in Fig.~\ref{fig:exp_gt_map}(d), the labels for \texttt{unlabeled} and \texttt{outlier} are represented in red points, and labels for \texttt{moving} objects are illustrated in purple. From the ground-truth static map~(Fig.~\ref{fig:exp_gt_map}(e)), we categorize terrain labels such as roads, terrain, and parking areas as traversable, while labeling objects like walls, fences, and poles as collision areas.
    However, to address the ambiguity in distinguishing between vegetation types such as tree leaves and grass, we exclude points labeled as \texttt{vegetation} with height over than $-1.0$. The rest of \texttt{vegetation} cells are designated as collision areas if $h_{\max} - h_{\min} > \tau^{h}$.

    \subsubsection{Evaluation Metrics}
    We evaluate terrain map reconstruction performance using the \textit{mean traversable terrain height error~(MTE)}, which specifically focuses on terrain relevant to walking and driving, where the ground-truth label indicates traversability, and the \textit{mean absolute height error~(MAE)}~\cite{yang2023neuraldense}.
    The MAE represents the height errors for the entire area and MTE measures height errors exclusively for traversable terrain.

    For navigation map performance regarding to collision cell decision, we leverage conventional metrics: \textit{precision}~($P$), \textit{recall}~($R$), $F_1$\textit{-score}~($F_1$), and \textit{accuracy}~($A$). 
    However, considering that the terrain map algorithms employ different criteria for dividing cells, we took into account whether there were collision areas in the adjacent regions when comparing it with $\mathcal{E}^{gt}$. 
    To determine the classification accuracy of the identified collision cells as true positives, we evaluated the performance based on the presence of adjacent ground-truth collision cells.

% \newpage
\section{Results and Discussion}

    % \vspace{-4mm}
    \subsection{Parameters}
    \begin{table}[ht!]
        % \vspace{-2mm}
        \captionsetup{font=footnotesize}
        \centering
        \caption{Parameter setting for terrain traversability mapping: Common used in comparison group are $\tau^{h}$ and $l$, while the other parameters are specific to TRIP. 
        Units of $h_{p}$, $l$, and $\tau^{h}$ are in m, whereas the others are unitless.}
        \setlength{\tabcolsep}{9.0pt}
        \begin{tabular}{l||c|c|c|c|c|c|c}
            \hline
            \multirow{2}{*}{Param.}     & \multirow{2}{*}{$\tau^{r}$}  & \multirow{2}{*}{$h_{p}$}  & \multirow{2}{*}{$\tau^{h}$}  & \multicolumn{2}{c|}{$l$}         & \multicolumn{2}{c}{$\tau^{\mathcal{M}}$} \\
            \cline{5-8} 
                                        &                                   &                           &                                   & \texttt{N} & \texttt{O}  & \texttt{S} & \texttt{D}     \\
            \hline
            Value       & 0.6             & 1.0         & 0.25           & 0.5    & 1.0                     & 3.0 & 1.0                \\
            \hline
        \end{tabular}
        \label{tab:param}
        % \vspace{-2mm}
    \end{table}
    As shown in Table~\ref{tab:param}, the parameters $l$ and $\tau^{\mathcal{M}}$ are set based on the environmental context, distinguishing between narrow~(\texttt{N}) or open~(\texttt{O}) spaces and static~(\texttt{S}) or dynamic~(\texttt{D}) conditions. It is noteworthy that the threshold $\tau^{h}$ is a common parameter utilized in baselines~\cite{shan2018bayesian},~\cite{wermelinger2016navigationplanning} as well.

    Moreover, the local terrain map completion range for comparison group was adjusted differently for each dataset: $6\text{m} \times 6\text{m}$ in the QRC environments~(\texttt{N}), $20\text{m} \times 20\text{m}$ for our campus dataset~(\texttt{O}), and $80\text{m} \times 80\text{m}$ in the SemanticKITTI~(\texttt{O}). These ranges were set depending one the robot's region of interest.
    The resolution of terrain map $\hat{\mathcal{E}}$ and ground-truth map $\mathcal{E}^{gt}$ are set as $0.2\text{m}$ for \texttt{O} and $0.1\text{m}$ for \texttt{N} environments.

    % \subsection{Quantitative results}
    \vspace{-3mm}
    \subsection{Result Analysis}
    For our evaluation, which encompasses terrain reconstruction and navigation mapping performances, we compared our proposed TRIP with several baselines: BGK$^{+}$\cite{shan2018bayesian}, which predicts BGK-based 2.5D terrain map with traversability; BGK$^{+}_{f}$, incorporating a box filter to eliminate outlier points; the robot-centric footprint traversability map~($t_{f}$-map)\cite{wermelinger2016navigationplanning}, derived from an elevation map\cite{fankhauser2014robot} with embedding traversability as a post-processing step; and TRIP-S, a simplified version of TRIP without the application of traversability-aware inference~(\ref{eq:tbgk_inference}) and outlier rejection~(\ref{eq:mahalanobis}).
    % Quantitative comparisons are presented in Table~\ref{tab:quantitative_comparison}, and qualitative terrain map results are depicted in Figs.~\ref{fig:qrc_result}~-~\ref{fig:kaist_result}. 
    Please note that $t_{f}$-map is suitable only for small-scale scenes, so it is only compared in QRC environments. 

    \begin{table}[t!]
        \captionsetup{font=footnotesize}
        \centering
        \caption{Quantitative comparison for the terrain map reconstruction and collision cell decision for navigation performance on the QRC simulation and the SemanticKIITI dataset. Units are cm for mean height error~(MHE) and mean traversable terrain error~(MTE), and \% otherwise.}
        \begin{tabular}{c|l||c|c|c|c|c|c}
            \hline
            \multicolumn{2}{c|}{\multirow{2}{*}{Metrics}}   &  \multicolumn{2}{c|}{Terrain Map $\downarrow$}           & \multicolumn{4}{c}{Navigation Map $\uparrow$}                     \\
            \multicolumn{2}{c|}{}                           &  MHE  & MTE  & $P_{~}$  & $R_{~}$  & $F_{1}$ & $A_{~}$ \\
            \hline
            \hline
            % Environment & \multicolumn{6}{c}{QRC Simulation Environment}\\
            % \hline
            \multirow{5}{*}{\rotatebox[origin=c]{90}{QRC Sim.}}         & BGK$^{+}$~\cite{shan2018bayesian}            & 22.31 & 21.38 & 46.1 & \textbf{99.2} & 62.9 & 75.3 \\
                                                                        & BGK$^{+}_{f}$                         & 18.60 & 17.42 & 56.6 & 99.1 & 72.0 & 84.2 \\
                                                                        & $t_{f}$-map~\cite{wermelinger2016navigationplanning}         & 14.19 & \textbf{~7.13} & 80.9 & 93.6 & 86.7 & 93.7 \\
                                                                        % &TRIP Light $d_{bgk}=0.5$                  & 12.56 & ~7.55 & 98.5 & 99.1 & 0.988 & 99.4 \\
                                                                        % &TRIP $d_{bgk}=0.5$                      & ~9.45 & ~6.61 & 99.6 & 98.7 & 0.992 & 99.6 \\
                                                                        % &TRIP Light $d_{bgk}=0.3$                  & 16.92 & 10.50 & 91.0 & 99.0 & 94.8 & 96.8 \\
                                                                        % &TRIP $d_{bgk}=0.3$                      & \textbf{10.17} & \textbf{~7.47} & \textbf{99.6} & 97.9 & \textbf{98.7} & \textbf{99.5} \\
                                                                        &TRIP-S                             & 16.92 & 10.50 & 91.0 & 99.0 & 94.8 & 96.8 \\
                                                                        &TRIP                         & \textbf{10.17} & ~7.47 & \textbf{99.6} & 97.9 & \textbf{98.7} & \textbf{99.5} \\
            \hline
            \hline
            % Environment & \multicolumn{6}{c}{KITTI, Seq.00 - \texttt{4300-4540}}\\
            % \hline
            \multirow{4}{*}{\rotatebox[origin=c]{90}{Seq.\texttt{00}}}         & BGK$^{+}$~\cite{shan2018bayesian}            & 31.34  & 20.56  & 64.6  &  93.7 & 76.5 & 81.2 \\
                                                                        & BGK$^{+}_{f}$                          & 29.83  & 18.35  & 70.6  &  93.7 & 80.5 & 85.3 \\
                                                                        % & TRIP Light $d_{bgk}=0.6$                  & 27.06  & 13.46  & 86.3  &  96.9 & 0.913 & 94.2 \\
                                                                        % & TRIP Light $d_{bgk}=1.0$                  & 28.35  & 14.37  & 81.6  &  96.7 & 0.885 & 92.1 \\
                                                                        % & TRIP $d_{bgk}=0.6$                      & 27.81  & 13.81  & 89.1  &  96.6 & 0.927 & 95.6 \\
                                                                        % & TRIP $d_{bgk}=1.0$                      & 28.05  & 13.77  & 88.0  &  96.8 & 0.922 & 95.3 \\
                                                                        & TRIP-S                             & 28.35  & 14.37  & 81.6  &  96.7 & 88.5 & 92.1 \\
                                                                        & TRIP                         & \textbf{28.05}  & \textbf{13.77}  & \textbf{88.0}  &  \textbf{96.8} & \textbf{92.2} & \textbf{95.3} \\
            \hline
            \hline
            % Environment & \multicolumn{6}{c}{KITTI, Seq.01 - \texttt{0000-0130}}\\
            % \hline
            \multirow{4}{*}{\rotatebox[origin=c]{90}{Seq.\texttt{01}}}         & BGK$^{+}$~\cite{shan2018bayesian}            & 21.81 & 17.73 & 68.9 & 93.3 & 79.3 & 87.4 \\
                                                                        & BGK$^{+}_{f}$                          & 21.48 & 17.27 & 72.7 & 93.2 & 81.7 & 89.1 \\
                                                                        % &TRIP Light $d_{bgk}=0.6$                  & 15.46 & 10.82 & 83.7 & 95.8 & 89.3 & 95.1 \\
                                                                        % &TRIP Light $d_{bgk}=1.0$                  & 16.85 & 12.15 & 81.7 & 94.1 & 87.5 & 94.4 \\
                                                                        % &TRIP $d_{bgk}=0.6$                      & 14.93 & 10.23 & 85.5 & 96.3 & 90.6 & 95.9 \\
                                                                        % &TRIP $d_{bgk}=1.0$                      & 15.51 & 10.59 & 85.4 & 95.5 & 90.2 & 95.8 \\
                                                                        & TRIP-S                             & 16.85 & 12.15 & 81.7 & 94.1 & 87.5 & 94.4 \\
                                                                        & TRIP                         & \textbf{15.51} & \textbf{10.59} & \textbf{85.4} & \textbf{95.5} & \textbf{90.2} & \textbf{95.8} \\
            \hline
            \hline
            % Environment & \multicolumn{6}{c}{KITTI, Seq.04 - \texttt{0000-0270}}\\
            % \hline
            \multirow{4}{*}{\rotatebox[origin=c]{90}{Seq.\texttt{04}}}         & BGK$^{+}$~\cite{shan2018bayesian}            & 31.10 & 25.68 & 62.8 & 81.4 & 70.9 & 79.4 \\
                                                                        & BGK$^{+}_{f}$                          & 30.82 & 25.24 & 64.4 & 81.4 & 71.9 & 80.4 \\
                                                                        % &TRIP Light $d_{bgk}=0.6$                  & 18.69 & 10.00 & 87.8 & 86.2 & 87.0 & 93.6 \\
                                                                        % &TRIP Light $d_{bgk}=1.0$                  & 20.86 & 11.14 & 84.5 & 82.7 & 83.6 & 91.3 \\
                                                                        % &TRIP $d_{bgk}=0.6$                      & 18.51 & ~9.69 & 90.0 & 87.6 & 88.7 & 95.1 \\
                                                                        % &TRIP $d_{bgk}=1.0$                      & 19.24 & ~9.98 & 89.1 & 86.9 & 88.0 & 94.7 \\
                                                                        & TRIP-S                                & 20.86 & 11.14 & 84.5 & 82.7 & 83.6 & 91.3 \\
                                                                        & TRIP                         & \textbf{19.24} & \textbf{~9.98} & \textbf{89.1} & \textbf{86.9} & \textbf{88.0} & \textbf{94.7} \\
            \hline
            \hline
            % Environment & \multicolumn{6}{c}{KITTI, Seq.05 - \texttt{2300-2760}}\\
            % \hline
            \multirow{4}{*}{\rotatebox[origin=c]{90}{Seq.\texttt{05}}}         & BGK$^{+}$~\cite{shan2018bayesian}            & 28.83 & 22.02 & 63.9 & 87.8 & 73.9 & 79.2\\
                                                                        & BGK$^{+}_{f}$                          & 27.75 & 20.40 & 69.1 & 87.8 & 77.3 & 82.7\\
                                                                        % &TRIP Light $d_{bgk}=0.6$                  & 21.60 & 10.64 & 86.8 & 93.1 & 89.8 & 93.3\\
                                                                        % &TRIP Light $d_{bgk}=1.0$                  & 24.05 & 12.24 & 81.1 & 92.1 & 86.2 & 90.4\\
                                                                        % &TRIP $d_{bgk}=0.6$                      & 22.03 & 10.66 & 91.8 & 92.1 & 91.9 & 95.2\\
                                                                        % &TRIP $d_{bgk}=1.0$                      & 22.89 & 11.04 & 90.0 & 92.1 & 91.0 & 94.7\\
                                                                        & TRIP-S                             & 24.05 & 12.24 & 81.1 & \textbf{92.1} & 86.2 & 90.4\\
                                                                        & TRIP                         & \textbf{22.89} & \textbf{11.04} & \textbf{90.0} & \textbf{92.1} & \textbf{91.0} & \textbf{94.7}\\
            \hline
            \hline
            % Environment & \multicolumn{6}{c}{KITTI, Seq.07 - \texttt{0600-0900}}\\
            % \hline
            \multirow{4}{*}{\rotatebox[origin=c]{90}{Seq.\texttt{07}}}         &BGK$^{+}$~\cite{shan2018bayesian}            & 38.77 & 34.79 & 63.1 & 90.0 & 74.2 & 74.8 \\
                                                                        &BGK$^{+}_{f}$                          & 37.72 & 33.12 & 65.4 & 89.9 & 75.7 & 76.9 \\
                                                                        % &TRIP Light $d_{bgk}=0.6$                  & 31.21 & 11.28 & 88.6 & 95.6 & 92.0 & 94 \\
                                                                        % &TRIP Light $d_{bgk}=1.0$                  & 33.67 & 12.82 & 83.8 & 94.8 & 89.0 & 91.3 \\
                                                                        % &TRIP $d_{bgk}=0.6$                      & 32.14 & 11.71 & 92.3 & 94.6 & 93.4 & 95.6 \\
                                                                        % &TRIP $d_{bgk}=1.0$                      & 32.95 & 11.87 & 91.5 & 94.5 & 93.0 & 95.3 \\
                                                                        &TRIP-S                             & 33.67 & 12.82 & 83.8 & \textbf{94.8} & 89.0 & 91.3 \\
                                                                        &TRIP                         & \textbf{32.95} & \textbf{11.87} & \textbf{91.5} & 94.5 & \textbf{93.0} & \textbf{95.3} \\
            \hline
        \end{tabular}
        \label{tab:quantitative_comparison}

    \end{table}
    \begin{table}[t!]
        % \vspace{-2mm}
        \captionsetup{font=footnotesize}
        \centering
        \caption{Mean processing time in ms for local estimation and global update on our real-world quadruped robot datasets using Intel(R) Core i7-8700 CPU.}
        \begin{tabular}{l||c|c|c|c|c|c}
            \hline
            Env.  & \multicolumn{3}{c|}{QRC London~(\texttt{N})}  & \multicolumn{3}{c}{Campus~(\texttt{O})} \\
            \hline
            Method                  & total  & local  & update                      & total & local & update     \\ %                      & local   & global  \\     
            \hline
            BGK$^{+}$               & 42.557 & ~4.783 & 37.774                      & 62.019 & ~2.107 & 59.912    \\ %                       & 9.00    & 149.825 \\
            BGK$^{+}_{f}$           & 43.344 & ~4.808 & 38.536                      & 58.046 & ~1.968 & 56.078    \\ %                       & 9.00    & 149.825 \\
            TRIP-S                  & 10.233 & ~9.295 & ~0.938                      & 14.787  & 13.789 & ~0.998    \\ %                       & 105.954 & 11.382 \\
            TRIP                    & ~\textbf{9.341} & ~8.656 & ~0.685             & \textbf{13.831}  & 13.337 & ~0.494    \\ %                       & 105.954 & 11.382 \\
            \hline
        \end{tabular}
        \label{tab:time}
        \vspace{-4mm}
    \end{table}
    \begin{figure}[t!]
        % \vspace{-2mm}
        \captionsetup{font=footnotesize}
        \centering
        \includegraphics[width=\columnwidth]{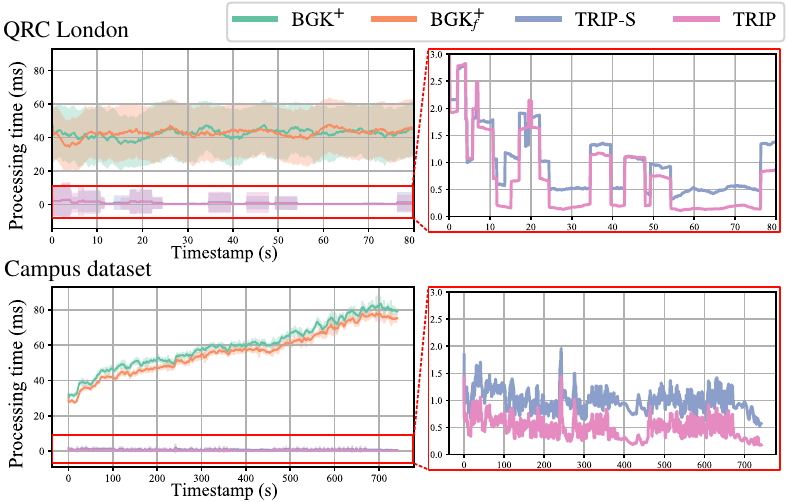}
        \caption{
                Global terrain terrain traversability map updating time comparison.
                }
        \label{fig:time_compare}
        \vspace{-4mm}
    \end{figure}

    \begin{figure*}[t!]
        \captionsetup{font=footnotesize}
        \centering
        \includegraphics[width=\textwidth]{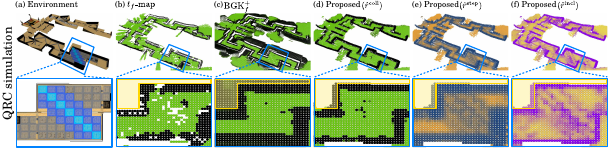}
        \includegraphics[width=\textwidth]{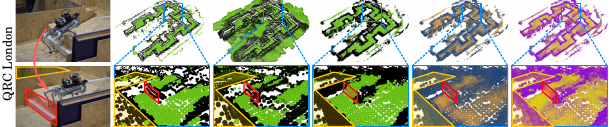}
        \caption{
                Qualitative results in Quadruped Robot Challenge~(QRC) environments:
                (a)~Example scenes featuring box stacks in simulated QRC and real QRC sites showcasing irregular terrains.
                (b)~$t_{f}$-map results highlighting empty cells in the terrain map after traversing the box stacks.
                (c)~BGK$^{+}_{f}$ results that predict non-observable areas~(highlighted as yellow areas), leading to inaccuracies in terrain representation.
                Our terrain map results with (d)~collision risk $\hat{r}^{\mathrm{coll}}$, (e)~inclination risk $\hat{r}^{\mathrm{incl}}$, and (f)~steppability risk $\hat{r}^{\mathrm{step}}$.
                The red boxes highlight that the baseline methods misidentified non-traversable areas on the narrow and irregular terrains.
                In contrast, our TRIP correctly identified the area as traversable, presenting detailed multi-modal risks.
                }
        \label{fig:qrc_result}
        \vspace{3mm}
        \includegraphics[width=\textwidth]{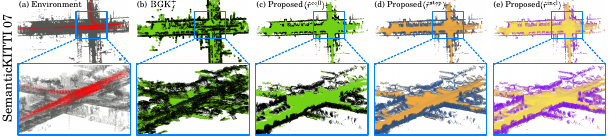}
        \caption{
                Qualitative results on SemanticKITTI dataset. 
                (a)~A naively accumulated 3d point cloud map of seq. \texttt{07} in the SemanticKITTI dataset, which consists of dynamic~(red) and static~(grey) points.
                (b)~BGK$^{+}_{f}$ results in terms of traversability.
                Our terrain map results comprised of (c)~collision risk~$\hat{r}^{\mathrm{coll}}$, (d)~inclination risk~$\hat{r}^{\mathrm{incl}}$, and (e)~steppability risk~$\hat{r}^{\mathrm{step}}$.
                While BGK$^{+}_{f}$ leaves the traces of moving objects, TRIP successfully rejects the effect of outliers, presenting clear risk maps.  
                % demonstrates robustness against these outliers.
                }
        \label{fig:kitti_result}
        \vspace{3mm}
        \includegraphics[width=\textwidth]{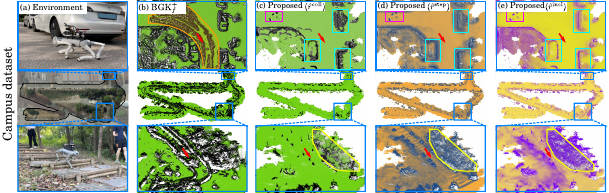}
        \caption{
                Qualitative results on our campus dataset:
                (a)~Example scenes depicting traversal over flat urban terrain and irregular stairs within a forest. 
                (b)~BGK$^{+}_{f}$ results in terms of traversability. 
                Our terrain map results comprised of (c)~collision risk~$\hat{r}^{\mathrm{coll}}$, (d)~inclination risk~$\hat{r}^{\mathrm{incl}}$, and (e)~steppability risk~$\hat{r}^{\mathrm{step}}$.
                The robot's positions are denoted by red arrows.
                BGK$^{+}_{f}$ is not only effected by outliers and overhanging elements but also fails to map traversability for various rough terrain types, such as stairs.
                TRIP offers discernible costs through a multi-modal traversability risk map, demonstrating its applicability in real environments.
                }
        \label{fig:kaist_result}
    \end{figure*}
   
    \subsubsection{Robustness In Environments of Various Scales}
    As shown in Table~\ref{tab:quantitative_comparison}, $t_{f}$-map has the lowest error in terms of MTE in the QRC simulation. 
    However, after navigating through the box-stacked region (blue grids in Fig.~\ref{fig:qrc_result}(a)), it leaves empty spaces~(Fig.~\ref{fig:qrc_result}(b)), which can pose challenges for subsequent navigation algorithms.
    
    In contrast, BGK$^{+}$ variants predict terrain without empty spaces in both narrow and wide environments, as shown in Figs.~\ref{fig:qrc_result}(c),~\ref{fig:kitti_result}(b), and~\ref{fig:kaist_result}(b). However, their tendencies to predict unobservable regions beyond the walls, which are emphasized as the yellow areas in Fig.~\ref{fig:qrc_result}(c), leads to lower performance, as presented in Table~\ref{tab:quantitative_comparison}.
    Particularly in narrow scenes, the erroneous inference of the baselines becomes more noticeable (the red regions in the bottom row of Figs.~\ref{fig:qrc_result}(b) and Figs.~\ref{fig:qrc_result}(c)), contributing to lower performance in terms of $P$, $F_1$, and $A$, despite high $R$.

    Evidenced by the highest $F_{1}$ and $A$ of TRIP across all test data, TRIP consistently provides a stable map for navigation regardless of the scale of surroundings. 
    Considering that there is no dynamic or sensor noise in the QRC simulation, the impact of Mahalanobis distance-based rejection is minimal. So, the results in the simulation show that T-BGK inference $\mathcal{L}^{\mathcal{T}}$ substantially improve both reconstruction and navigation.
    % Considering the minimal impact of our outlier rejection in a static environment like QRC simulation, when compared with TRIP-S, T-BGK can substantially improve both reconstruction and navigation.

    \subsubsection{Robustness Against Outliers}
    As shown in Figs.~\ref{fig:kitti_result} and~\ref{fig:kaist_result}, BGK$^{+}_f$ failed to reject outliers, including sensor noise and dynamic objects, leading to higher MHE and MTE in the SemanticKITTI sequences, as reported in Table~\ref{tab:quantitative_comparison}.
    Particularly notable is the comparison with TRIP-S results,
    highlighting that TRIP exhibits less impact from dynamic objects and outliers. 
    This supports our key claim that the steppability-based Mahalanobis distance filtering enhances the terrain map reconstruction quality.

    \subsubsection{Multi-Modal Traversability Risk Map}
    Our map results employ not only $r^{\mathrm{coll}}$ but also $r^{\mathrm{incl}}$, which indicates risky areas within each zone, such as box stacks in the second row of Fig.~\ref{fig:qrc_result}, and guarantees safety in regions distant from wall-like vertical obstacles. Additionally, $r^{\mathrm{step}}$ acts as an indicator of the risk posed by terrain edges or irregular terrains.
    As illustrated by the yellow polygons in Figs.~\ref{fig:kaist_result}(c)-(e), $r^{\mathrm{incl}}$ and $r^{\mathrm{step}}$ layers collectively highlight high-risk situations when $r^{\mathrm{coll}}$ incorrectly estimates an area, e.g. vine forest, as safe. This information can be instrumental in optimizing the navigation algorithm.

% \newpage
\section{Conclusion and Future works}
    \vspace{-1mm}
    TRIP significantly enhances the online navigation system by displaying multi-modal normalized traversability risks, especially on irregular terrains. 
    Our experiments highlight the following key capabilities:
    a)~Utilizing spherical coordinates, we achieve scalable terrain property estimation, addressing sparsity issues across various environments - from narrow to open and wild environment.
    b)~T-BGK inference, coupled with $r^{\mathrm{step}}$-based prediction, enhances local terrain map completion by emphasizing three traversability risks.
    c)~TRIP demonstrates robustness against outliers from dynamic elements and noise via steppability-based Mahalanobis distance filtering.
    
    However, our terrain traversability map lacks instance segmentation updates, limiting determinations of collision and hazardous zones for objects, e.g. cyan boxes in Figs.\ref{fig:kaist_result}(c)-(e). Initial processing of nearby dynamic objects also poses challenges, e.g. purple boxes in Figs.\ref{fig:kaist_result}(c)-(e).
    In our future work, we aim to integrate instance-aware updates into TRIP for handling more complex navigation scenarios.
    \vspace{-2mm}
\bibliographystyle{IEEEtran}
\bibliography{./ref}

\end{document}